\documentclass[journal,9pt]{IEEEtran}
\usepackage[latin9]{inputenc}
\usepackage{amsmath}
\usepackage{amssymb}
\usepackage{graphicx}

\makeatletter

\providecommand{\tabularnewline}{\\}

\ifCLASSINFOpdf
\else
\fi
\makeatother

\begin{document}

\title{Category-Based Deep CCA for Fine-Grained Venue Discovery from Multimodal
Data}

\author{\IEEEauthorblockN{Yi Yu$^{1}$, Suhua Tang$^{2}$, Kiyoharu Aizawa$^{3}$, Akiko Aizawa$^{1}$}
\\
 \IEEEauthorblockA{$^{1}$Digital Content and Media Sciences Research Division, National
Institute of Informatics \\
 $^{2}$Graduate School of Informatics and Engineering, The University
of Electro-Communications \\
 $^{3}$Dept. of Information and Communication Engineering, The University
of Tokyo\\
 }}
\maketitle
\begin{abstract}
In this work, travel destination and business location are taken as
venues. Discovering a venue by a photo is very important for context-aware
applications. Unfortunately, few efforts paid attention to  complicated
real images such as venue photos generated by users. Our goal is
fine-grained venue discovery from heterogeneous social multimodal
data. To this end, we propose a novel deep learning model, Category-based
Deep Canonical Correlation Analysis (C-DCCA). Given a photo as input,
this model performs (i) exact venue search (find the venue where the
photo was taken), and (ii) group venue search (find relevant venues
with the same category as that of the photo), by the cross-modal correlation
between the input photo and textual description of venues. In this
model, data in different modalities are projected to a same space
via deep networks. Pairwise correlation (between different modal data
from the same venue) for exact venue search and category-based correlation
(between different modal data from different venues with the same
category) for group venue search are jointly optimized. Because a
photo cannot fully reflect rich text description of a venue, the number
of photos per venue in the training phase is increased to capture
more aspects of a venue. We  build a new venue-aware multimodal dataset
by integrating Wikipedia featured articles and Foursquare venue photos.
Experimental results on this dataset confirm the feasibility of the
proposed method. Moreover, the evaluation over another publicly available
dataset confirms that the proposed method outperforms state-of-the-arts
for cross-modal retrieval between image and text.
\end{abstract}


\section{Introduction}

Context aware applications are very promising because they can provide
suitable services adapted to user context. Assume a user visits a
venue for the first time. He does not know exactly where he is, but
takes a photo there. Venue discovery helps to find the exact venue
where the photo was taken and a group of relevant venues whose textual/visual
features also match the photo. The former captures user context while
the latter is important for venue recommendation.

Fine-grained venue discovery from a photo was almost impossible years
ago since we lack reliable venue data sources. Fostered by multimedia
technology innovation and mobile user engagements, business-related
social multimedia data and information have been expanding with large
volumes over the Internet, e.g., featured articles for business venues
in Wikipedia, business venue photos on Foursquare and Yelp, and video
advertising on YouTube.  On the other hand, the growth of venue photos
containing visual business contents has made various business services
significantly more visible to a seeker on the Internet, leading to
a visit or a purchase in the physical world. The interactions between
users and venues result in various multimedia data and information
aggregated on the Internet, which brings us new opportunities to fine-grained
venue discovery by leveraging the power of social multimodal data.
Here, ``multimodal'' means that each venue has multiple representations
in different modalities like text and vision (images).

Some literature has investigated venue discovery, e.g., the prediction
of geographic category of a photo \cite{Hays08} or visual concept
of a photo \cite{Chen16}, or the coarse prediction of the location
of a photo \cite{Schindler07}. However, few efforts focus on fine-grained
venue discovery with more complicated real images generated by users
such as venue photos containing objects, geographic categories, and
more meaningful semantic descriptions. 

Fine-grained venue discovery relies on the correlation analysis between
images and text description of venues. This is a ``cross-modal''
analysis because it compares data in one modality with data in a different
modality. Many efforts have been carried out for single modal or cross-modal
classification and retrieval tasks. Researches in a single-modal task
include, e.g., using deep networks for scene recognition \cite{Zhou14},
using deeper networks to achieve better performance in image classification
\cite{Simonyan14,Szegedy15}, generalizing features extracted from
a specific dataset in a fully supervised fashion for generic tasks
\cite{Donahue14}. Some works investigated multiple modalities, e.g.,
event summarization by leveraging both visual and textual information
\cite{Rajiv16}. Recently, there are some initial work on cross-modal
correlation, such as predicting answers given an image and a question
as input \cite{Malinowski15}, analyzing pairwise correlation between
images and their captions \cite{Yan15}, cross-modal retrieval by
canonical correlation analysis (CCA) \cite{Rasiwasia10} or kernel
CCA (KCCA) \cite{Rasiwasia14}.

In this work, we investigate venue-related multimodal data from Wikipedia
and Foursquare, and study (i) exact venue search (find the venue where
the photo was taken), and (ii) group venue search (find relevant venues
with the same category as that of the photo) in a joint framework
for fine-grained venue discovery. To the best of our knowledge, this
is the first study that focuses on the joint optimization by visual
and textual diversity of venues over an integrated venue-related multimodal
data. To this end, we propose a Category-based Deep CCA (C-DCCA) method,
where data in different modalities are non-linearly projected to the
same space via deep networks so that data of different modalities
from the same venue or from different venues with the same category
are highly correlated in that space. 

The contribution of this paper is three-fold, as follows: (i) The
proposed C-DCCA method is an important extension to Deep CCA (DCCA)
\cite{Andrew13}. Pairwise correlation between different modal data
from the same venue and category-based correlation between different
modal data from different venues with the same category are jointly
optimized so that a rank list of relevant venues with the same category
as that of the input photo is predicted, and the exact venue where
the photo was taken appears in the top. (ii) Fine-grained venue discovery
is done by leveraging heterogeneous multimodal contents. Although
each featured article of a venue in Wikipedia contains rich text information
about the venue, its accompanying photo cannot well represent all
the visual information. Therefore, we further exploit venue photos
from Foursquare. In this way, the number of photos per venue in the
training phase is increased to capture rich text information of a
venue. (iii) Extensive experiments on multimodal venue-related dataset
verify the feasibility of the proposed C-DCCA model. Additionally,
the evaluation on a publicly available dataset \cite{Rasiwasia10}
confirms that the proposed method outperforms state-of-the-arts for
cross-modal retrieval between image and text. 

The rest of this paper is organized as follows. Related work is reviewed
in Sec.~\ref{sec:Related}. Sec.~\ref{sec:Preliminary} explains
the preliminary of CCA and DCCA. Then, Sec.~\ref{sec:Algorithm}
presents the proposed C-DCCA architecture in our work. Experimental
evaluation results are shown in Sec.~\ref{sec:Evaluation}. Finally,
Sec.~\ref{sec:Conclusion} concludes this paper.

\section{Related work}

\label{sec:Related}Generally speaking, fine-grained venue discovery
by leveraging heterogeneous social multimodal dataset is a very challenging
research topic. It is related to two research lines in the following.

\subsection{Cross modal correlation learning}

Automatically learning to represent cross-modal correlations between
images and texts has attracted lots of research interests in the areas
of computer vision, natural language processing, and machine learning.
CCA \cite{Hotelling36}, cross-modal factor analysis (CFA) \cite{Li03},
Bilinear model \cite{Tenenbaum00} as joint dimensionality reduction
techniques were proposed to learn linear functions that map different
modalities to the same canonical (semantic) space where the correlation
between two modalities is maximized. It is a very popular embedding
method for matching images against texts. KCCA \cite{Cristianini00}
is an extension of CCA to calculate non-linear correlations by leveraging
the kernel trick. In comparison, DCCA \cite{Andrew13} models non-linear
mapping by deep neural networks and on this basis maximizes correlation
across different modalities. Cross-modal correlation between images
and their captions via DCCA is analyzed in \cite{Yan15}. General
cross-modal correlation between images and rich text information is
studied in \cite{Rasiwasia10} by CCA and in \cite{Rasiwasia14} by
KCCA. 

\subsection{Predicting venues from images}

There are some existing works which concentrate on location or venue
category prediction from images \cite{Hays08,Chen16,Schindler07,Friedland10,Li09,Chen11,Lin15}.
% Unfortunately, the granularity of geo-tagged% image location is coarse in a city level, and is mainly suitable for% outdoor images.
Earliest research \cite{Hays08} related to inferring geo-information
from an image is to estimate a distribution of geographic locations
by utilizing a data-driven scene matching method. However, it targets
for outdoor images at the level of coarse granularity such as a city.
Authors in \cite{Schindler07} proposed to select the vocabulary by
using the most informative features and exploit vocabulary tree search
for large-scale location discovery within a city based on investigating
street view images. The level of granularity is within a 20 km urban
terrain. In \cite{Friedland10}, the authors leverage multiple sources
of information to infer the location of a video. Their location discovery
is at region-scale granularity. Moreover, the task of landmark discovery
\cite{Li09,Chen11} is to recognize buildings or objects from a given
image, which is closely related to location prediction. It usually
needs to classify images according to objects. Some hand-crafted features
such as SIFT are extracted to represent images. For a more recent
work in \cite{Lin15}, the authors learn deep representations to localize
a ground-level image where aerial images are used to help location
prediction. They are able to obtain a finer granularity but only applicable
to outdoor images with buildings in a city scale. Some researchers
have tried to use CNN (Convolutional neural network) to detect visual
concepts as textual words, convert them to a word vector, and combine
this with visual feature to represent venues for venue recognition
\cite{Chen16}. This, however, in essence used only visual information.

\subsection{A short comparison}

Distinguished from state-of-the-art approaches that mainly focus on
pairwise correlation \cite{Rasiwasia10,Yan15} (between different
view data of the same object), we introduce category information into
the two-branch deep network.  In contrast, we optimize both pairwise
correlation and category-based correlation in a joint framework. Compared
with the methods that predict coarse location from images at city-scale
granularity \cite{Schindler07,Friedland10}, fine-grained venue discovery
in our work aims to search a relevant venue, the context of a photo
indicating where the photo was taken, by leveraging the cross-modal
correlation between images and rich text information of venues. In
addition, we use more photo resources to better capture different
visual aspects of a venue.

\section{Preliminaries}

\label{sec:Preliminary}CCA has been a very popular method for embedding
multimodal data in the shared space. Before presenting our C-DCCA
model, we first give an outline of CCA and its typical extensions:
KCCA and DCCA.

Let $\boldsymbol{x}\in R^{m}$ (e.g., visual feature) and $\boldsymbol{y}\in R^{n}$
(e.g., textual feature) be zero mean random (column) vectors with
covariances $\boldsymbol{C}_{xx}$, $\boldsymbol{C}_{yy}$ and cross-covariance
$\boldsymbol{C}_{xy}$. When a linear projection is performed, CCA
\cite{Hotelling36} tries to find two canonical weights $\boldsymbol{w}_{x}$
and $\boldsymbol{w}_{y}$, so that the correlation between the linear
projections $u=\boldsymbol{w}_{x}^{T}\boldsymbol{x}$ and $v=\boldsymbol{w}_{y}^{T}\boldsymbol{y}$
is maximized.

\begin{eqnarray}
(\boldsymbol{w}_{x},\boldsymbol{w}_{y}) & = & \underset{(\boldsymbol{w}_{x},\boldsymbol{w}_{y})}{argmax}\:corr(\boldsymbol{w}_{x}^{T}\boldsymbol{x},\boldsymbol{w}_{y}^{T}\boldsymbol{y})\nonumber \\
 & = & \underset{(\boldsymbol{w}_{x},\boldsymbol{w}_{y})}{argmax}\frac{{\boldsymbol{w}_{x}^{T}\boldsymbol{C}_{xy}\boldsymbol{w}_{y}}}{\sqrt{\boldsymbol{w}_{x}^{T}\boldsymbol{C}_{xx}\boldsymbol{w}_{x}\cdot\boldsymbol{w}_{y}^{T}\boldsymbol{C}_{yy}\boldsymbol{w}_{y}}}.\label{eq:CCA-vec}
\end{eqnarray}
When multiple pairs of canonical weights are needed, it is equivalent
to solve the following equation

\begin{eqnarray}
(\boldsymbol{W}_{x},\boldsymbol{W}_{y}) & \negthinspace\negthinspace\negthinspace\negthinspace=\negthinspace\negthinspace\negthinspace\negthinspace & \underset{(\boldsymbol{W}_{x},\boldsymbol{W}_{y})}{argmax}\:tr(\boldsymbol{W}_{x}^{T}\boldsymbol{C}_{xy}\boldsymbol{W}_{y})\nonumber \\
\text{{subject\,to}}: & \negthinspace\negthinspace\negthinspace\negthinspace\negthinspace\negthinspace\negthinspace\negthinspace & \negthinspace\negthinspace\negthinspace\negthinspace\negthinspace\negthinspace\boldsymbol{W}_{x}^{T}\boldsymbol{C}_{xx}\boldsymbol{W}_{x}=\boldsymbol{I,}\:\boldsymbol{W}_{y}^{T}\boldsymbol{C}_{yy}\boldsymbol{W}_{y}=\boldsymbol{I},\label{eq:CCA-mat}
\end{eqnarray}
where $\boldsymbol{I}$ is an identity matrix, $tr(\cdot)$ is the
trace of a matrix, $\boldsymbol{W}_{x}$ and $\boldsymbol{W_{y}}$
are weight matrices with $\boldsymbol{w}_{x}$ and $\boldsymbol{w}_{y}$
as their columns. With $\boldsymbol{T}\triangleq\boldsymbol{C}_{xx}^{-1/2}\boldsymbol{C}_{xy}\boldsymbol{C}_{yy}^{-1/2}$
and let $\boldsymbol{U}_{k}$ and $\boldsymbol{V}_{k}$ be the first
$k$ left- and right- singular vectors of $\boldsymbol{T}$, $\boldsymbol{W}_{x}$
and $\boldsymbol{W_{y}}$ can be computed as follows

\begin{equation}
\boldsymbol{W_{x}}=\boldsymbol{C}_{xx}^{-1/2}\boldsymbol{U}_{k},\boldsymbol{W_{y}}=\boldsymbol{C}_{yy}^{-1/2}\boldsymbol{V}_{k}.\label{eq:CCA-sol}
\end{equation}
In the actual computation, $\boldsymbol{C}_{xx}$ and $\boldsymbol{C}_{yy}$
are estimated from instances of $\boldsymbol{x}$ and $\boldsymbol{y}$
with regularization. Once the canonical weights ($\boldsymbol{W}_{x}$
and $\boldsymbol{W_{y}}$) are learned, they can be used to map new
data of different modalities to the same space for correlation analysis.
One of the known shortcomings of CCA is that its linear projection
may not well model the nonlinear relation between different modalities.

KCCA \cite{Cristianini00} extends CCA to calculate non-linear correlations,
and tries to model the nonlinearity of low dimensional space as a
linear problem in a high dimensional space. Assume $\phi_{x}$ maps
$\boldsymbol{x}\in R^{m}$ as $\phi_{x}(\boldsymbol{x})\in R^{N}$
($N>m$) and $\phi_{y}$ maps $\boldsymbol{y}\in R^{n}$ as $\phi_{y}(\boldsymbol{y})\in R^{N}$
($N>n$). With $p$ pairs of samples $\boldsymbol{X}=[\boldsymbol{x}_{1},\boldsymbol{x}_{2},...,\boldsymbol{x}_{p}]$
and $\boldsymbol{Y}=[\boldsymbol{y}_{1},\boldsymbol{y}_{2},...,\boldsymbol{y}_{p}]$,
$p$ pairs of data in the logical high dimensional space can be obtained
as $\phi_{\boldsymbol{X}}=[\phi_{x}(\boldsymbol{x}_{1}),\phi_{x}(\boldsymbol{x}_{2}),...,\phi_{x}(\boldsymbol{x}_{p})]$
and $\phi_{\boldsymbol{Y}}=[\phi_{y}(\boldsymbol{y}_{1}),\phi_{y}(\boldsymbol{y}_{2}),...,\phi_{y}(\boldsymbol{y}_{p})]$.
Then, with $u=\boldsymbol{w}_{x}^{T}\phi_{\boldsymbol{X}}$ and $v=\boldsymbol{w}_{y}^{T}\phi_{\boldsymbol{Y}}$,
the basic CCA is applied as follows:

\begin{equation}
(\boldsymbol{w}_{x},\boldsymbol{w}_{y})=\underset{(\boldsymbol{w}_{x},\boldsymbol{w}_{y})}{argmax}\:corr(\boldsymbol{w}_{x}^{T}\phi_{\boldsymbol{X}},\boldsymbol{w}_{y}^{T}\phi_{\boldsymbol{Y}}).\label{eq:KCCA-def}
\end{equation}
$\boldsymbol{w}_{x}$ and $\boldsymbol{w}_{y}$ themselves can be
represented by $\phi_{\boldsymbol{X}}$ and $\phi_{\boldsymbol{Y}}$
as $\boldsymbol{w}_{x}=\phi_{\boldsymbol{X}}\boldsymbol{\alpha}_{x}$
and $\boldsymbol{w}_{y}=\phi_{\boldsymbol{Y}}\boldsymbol{\alpha}_{y}$,
and the above equation can be rewritten as

\begin{eqnarray}
\negthinspace\negthinspace\negthinspace\negthinspace(\negthinspace\boldsymbol{\alpha}_{x},\boldsymbol{\alpha}_{y}\negthinspace) & \negthinspace\negthinspace\negthinspace=\negthinspace\negthinspace\negthinspace & \underset{(\boldsymbol{\alpha}_{x},\boldsymbol{\alpha}_{y})}{argmax}\frac{{\boldsymbol{\alpha}_{x}^{T}K_{\boldsymbol{X}}K_{\boldsymbol{Y}}\boldsymbol{\alpha}_{y}}}{\sqrt{\boldsymbol{\alpha}_{x}^{T}K_{\boldsymbol{X}}^{2}\boldsymbol{\alpha}_{x}\cdot\boldsymbol{\alpha}_{y}^{T}K_{\boldsymbol{Y}}^{2}\boldsymbol{\alpha}_{y}}}\label{eq:KCCA}
\end{eqnarray}
where $K_{\boldsymbol{X}}=\phi_{\boldsymbol{X}}^{T}\phi_{\boldsymbol{X}}$
and $K_{\boldsymbol{Y}}=\phi_{\boldsymbol{Y}}^{T}\phi_{\boldsymbol{Y}}$
are defined as kernels. Similar to CCA, KCCA also maps $\boldsymbol{x}$
and $\boldsymbol{y}$ to the same canonical (kernel) space, although
non-linearly. These kernels and canonical weights obtained in the
training stage can be used to map new data into the canonical space
for correlation analysis. One of the potential problems of KCCA is
that it may overfit to the training data.

DCCA \cite{Andrew13} also tries to calculate non-linear correlations
between different modalities by a combination of DNNs (deep neural
networks) and CCA. Different from KCCA which relies on kernel functions
(corresponding to a logical high dimensional (sparse) space), DNN
has the extra capability of compressing features to a low dimensional
(dense) space, and then CCA is implemented in the objective function.
The DNNs, which realize the non-linear mapping ($\varphi_{x}(\cdot)$
and $\varphi_{y}(\cdot)$), and the canonical weights ($\boldsymbol{w}_{x}$
and $\boldsymbol{w}_{y}$), are trained to maximize the correlation
after the non-linear mapping, as follows.

\begin{equation}
(\boldsymbol{w}_{x},\boldsymbol{w}_{y},\varphi_{x},\varphi_{y})\negthinspace=\negthinspace\negthinspace\underset{(\boldsymbol{w}_{x}\negthinspace,\boldsymbol{w}_{y}\negthinspace,\varphi_{x}\negthinspace,\varphi_{y})}{argmax}\negthinspace\negthinspace corr(\boldsymbol{w}_{x}^{T}\varphi_{x}(\boldsymbol{x}),\boldsymbol{w}_{y}^{T}\varphi_{y}(\boldsymbol{y})).
\end{equation}

\section{Algorithm}

\label{sec:Algorithm} 

Here, we consider two tasks in the same framework: (i) exact venue
search (find the venue where the photo was taken), and (ii) group
venue search (find relevant venues with the same category as that
of the photo) for a given photo without accurate location information\footnote{1Although GPS metadata is available for most outdoor photos, its accuracy
may be low in urban canyons with high buildings. In addition, it is
not available for indoor or underground environments. Therefore, we
consider GPS metadata as an option, not a necessity.}, by the correlation between the input photo and textual descriptions
of venues in the database. Each venue has an assigned category, rich
text description and user generated images. Text and images inherit
the same category of a venue. In the training phase, text and images
of venues are used to learn the cross-modal correlation so that text
features and visual features are highly correlated in the canonical
space and text feature alone can well represent a venue. In the testing
phase, visual feature of an input photo is compared against the textual
features of venues. In the future, we will further explore how to
use extra images to better represent a venue. 

Typically Wikipedia data contains one featured article and one photo
per venue. However, a single photo cannot well represent all visual
aspects of a venue. In contrast, Foursquare contains many photos generated
by users for each venue.  In the training phase, we try to use Foursquare
photos to represent diverse visual aspects of a specific venue, described
in the Wikipedia featured article. In other words, multiple Foursquare
photos are used to represent the same Wikipedia article, and featured
articles in Wikipedia and Foursquare photos for the same venues are
integrated together so as to better learn the correlation between
images and text.

Contrary to the existing methods of visual venue discovery, we propose
category-based deep canonical correlation analysis (C-DCCA) for the
cross-modal search between images and text. Our approach is an important
extension of DCCA, which handles the correlation learning of diverse
visual and textual contents for the specific venues in the real scenario
of multimodal learning.

In the following, we express the motivation of learning strategy,
formulate the problem of fine-grained venue discovery, and present
the architecture of the proposed deep network in details. 

\subsection{Motivation of learning strategy}

The ability to infer correlations between images and texts of venues
is a prerequisite for the prediction of relevant venues for a given
photo. As discussed in previous sections, today we can acquire hundreds
of venue photos or even more for a specific venue with different venue
aspects in various media sharing platforms. Existing cross-modal correlation
learning is a flat method which predicts the correlation based on
the pairwise similarity between an image and a text article, which
is not sufficient for learning the diverse aspects of a specific venue
and the correlation between image and text of different venues of
the same category. In this work, we make use of the Wikipedia featured
article for a specific venue as its knowledge base and the corresponding
venue photos in Wikipedia and Foursquare as its visual contents, if
the Foursquare image is visually similar to the Wikipedia image and
is explicitly described by Wikipedia description. % (e.g. the venue is a restaurant, the Foursquare image is an exterior% wall whose architectural features are definitely related to the outside% Wikipedia image of the venue, and naturally we are aware that this% is a venue for the restaurant due to a sign of a menu or a restaurant% or clients and tables).
In this way, each venue has multimodal data (geographic categories,
visual contents, textual descriptions), and needs to be represented
in the same space for the similarity comparison.

In many deep learning methods for recognition and discovery issues,
fine-tuning layers suitable for new tasks is non-trivial. Moreover,
some architectures are developed to jointly consider two or more tasks
by utilizing different branches to concatenate each other. In this
study, we tackle fine-grained venue discovery by optimizing the CCA
objective function in the joint deep learning framework. Besides the
pairwise correlation considered in DCCA, we also try to learn the
category-based correlation between data of different venues with the
same category.

\subsection{Problem formulation}

As for a venue $V$, we denote its Wikipedia text as $\boldsymbol{t}{}_{W,V}$,
its Wikipedia photo as $\boldsymbol{i}{}_{W,V}$, and its Foursquare
photo set as $\boldsymbol{I}{}_{F,V}$. Given a social photo $\boldsymbol{i}$,
the system will predict a rank list of venues, which includes the
exact venue where the photo was taken and relevant venues with the
same category as that of the photo, based on the correlation between
the input photo and textual description of venues

\begin{equation}
RankList(V)=\underset{V}{sort}\:corr(\boldsymbol{i},\boldsymbol{t}_{W,V}).
\end{equation}

Because deep features from image and featured articles of venues are
mapped into a common space where they are highly correlated, here
textual feature alone is used to represent a venue. The function $corr(\cdot)$,
which takes into account both the pairwise correlation among data
from the same venue and the category-based correlation among data
from different venues with the same category, is defined later by
C-DCCA.

To improve the accuracy of predicting the category and relevant venues
of a photo, in this work, $\boldsymbol{I}{}_{F,V}$ is divided into
two parts: $\boldsymbol{I}{}_{F,V,1}$ is used together with $\boldsymbol{i}{}_{W,V}$
to train the cross-modal model, and $\boldsymbol{I}{}_{F,V,2}$ is
used for the test. 

\subsection{Architecture of the proposed network}

\begin{figure*}
\centering

\includegraphics[width=16cm]{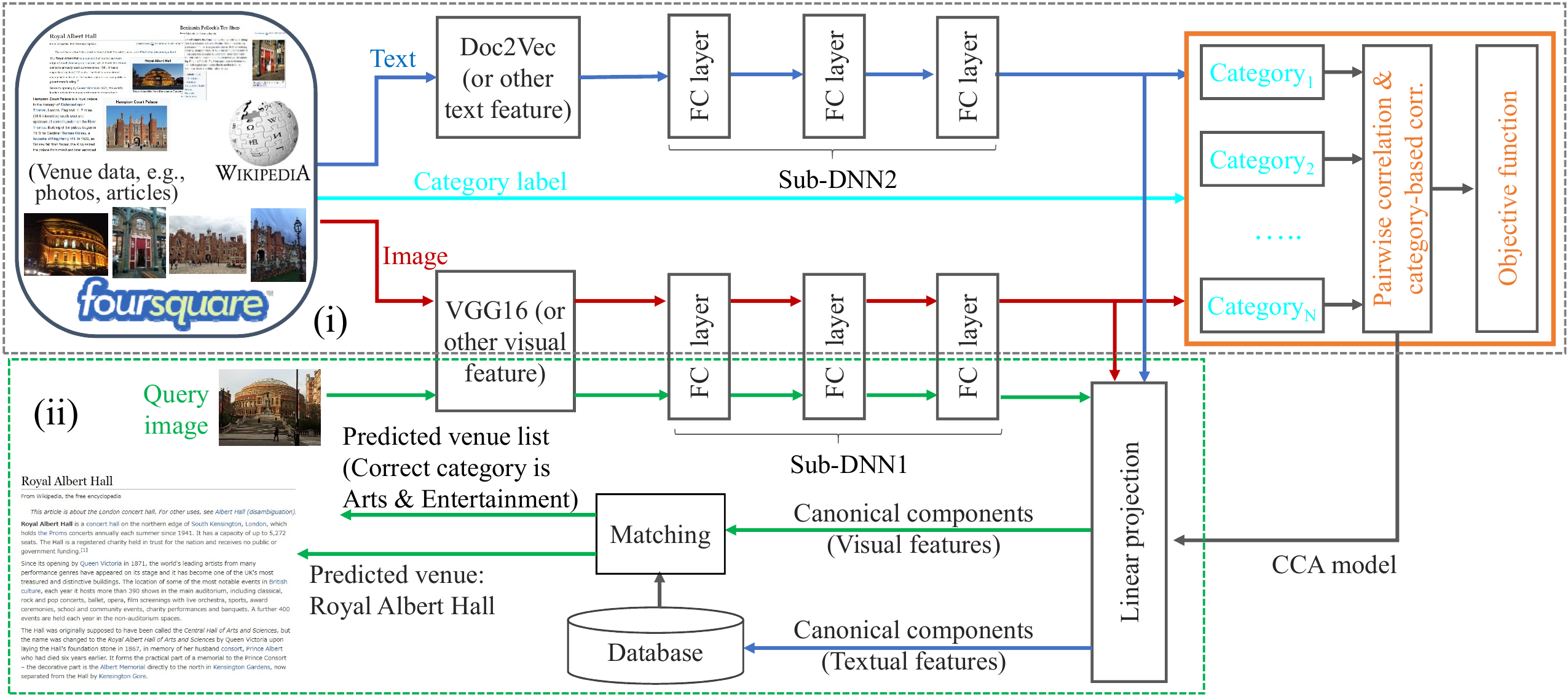}

\vspace{-2mm}

\caption{\label{fig:framework} (i) Proposed C-DCCA architecture. From images
and text articles of venues in the database, visual features and textual
features are extracted (e.g., by using VGG16 and Doc2Vec, respectively).
These features are further mapped to a common space, by using sub-DNN1
and sub-DNN2. Then, in the common space, new features of images and
texts are divided into groups according to their categories, and CCA
is applied in the objective function to adjust sub-DNN1/2, considering
both pairwise correlation between features from the same venue, and
category-based correlation between features from different venues
with the same category. (ii) Venue discovery. With a given photo as
input, a rank list, which contains the exact venue where the photo
was taken and a group of relevant venues with the same category as
that of the input photo, is predicted.}

\vspace{-1mm}
\end{figure*}

Figure~\ref{fig:framework} shows the whole framework. Part (i) in
dashed, gray line illustrates the proposed network architecture. From
the images and text articles, visual features and textual features
are extracted respectively. These features, however, belong to different
modalities and cannot be compared directly. Therefore, they are mapped
to a common space, by using sub-DNNs. To enhance the correlation in
this common space, CCA is used as the objective function. Either of
the sub-DNN models consists of three fully connected layers. To capture
the common features of similar venues, we use geographic categories
defined in Foursquare as concepts to divide venues into groups. Both
the pairwise correlation between photos and text of the same venue,
and the category-based correlation between photos and text of different
venues with the same category, are considered in the CCA objective
function, which is used to adjust the sub-DNNs.

\subsubsection{Visual feature extraction}

Recently CNNs have demonstrated excellent performances in image recognition
tasks. In particular, CNNs have the great capability in learning complex
features for representing visual contents, which is superior to hand-crafted
features such as HOG \cite{Dalal05} and SIFT \cite{Lowe99}. Therefore,
in the image branch, we take the VGG16 model \cite{Simonyan14} pre-trained
on ImageNet as an example to extract visual feature for all the images,
although other visual feature extraction method is also applicable. 

Each venue image is first converted to a fixed size of $224\times224$,
and then input into the network. The VGG16 model consists of 13 convolutional
layers (conv1-conv13) and three fully connected layers (fc14-fc16).
All layers use a ReLU (rectified linear unit) activation except fc16
which uses a softmax activation for the purpose of image classification.
 Each fully connected layer, except the last one, is followed by
a dropout layer, to avoid overfitting. Images are processed sequentially
per layer, and finally the 4,096-dimensional feature of fc15 is extracted
as the visual feature for each venue image.

\subsubsection{Textual feature extraction}

Many features have been proposed for representing text articles, e.g.,
TF-IDF (term frequency-inverse document frequency), topic models such
as latent Dirichlet allocation (LDA) model, and word embedding method
that represents each word by a vector in a space (Word2Vec) where
words with similar meaning are close to each other in the space. Doc2Vec
\cite{Lau16} extends the Word2Vec model by converting an entire document
into a fixed length vector, taking into account the order of words
in the context.

In the text branch, we take the Doc2Vec model as an example to extract
textual feature. Text description of each venue, crawled from Wikipedia,
consists of much non-relevant information. First, only the main text
and category information are extracted by invoking Wikipedia API.
Then, it is tokenized by using coreNLP \cite{ManningSBFBM14}, and
passed to  the Doc2Vec model, generating a fixed 300-dimensional
feature for each venue article. We use the pre-trained apnews\_dbow
weights\footnote{https://ibm.ent.box.com/s/9ebs3c759qqo1d8i7ed323i6shv2js7e}
in the analysis.

\subsubsection{Category-based DCCA}

\label{sec:C-DCCA}Visual features and textual features are further
converted into low dimensional features in a common space by using
different sub-DNNs. The details of sub-DNNs are shown in Table~\ref{tab:sub-dnn}.
These two sub-DNNs each have 3 fully connected layers. Before the
input, there is a batch normalization. In the 1st layer and 2nd layer,
there is a dropout sub-layer, used for avoiding over-fitting. Each
layer takes the output of its preceding layer $\boldsymbol{d}_{i-1}$
to compute its output $\boldsymbol{d}_{i}=f_{i}(\boldsymbol{\Psi}_{i}\boldsymbol{d}_{i-1}+\boldsymbol{b}_{i})$,
where $\boldsymbol{\Psi}_{i}$ and $\boldsymbol{b}_{i}$ are the weight
matrix and bias for the $i$th layer and $f_{i}(\cdot)$ is the activation
function.

\begin{table}[!t]
\caption{Structure of sub-DNNs}

\label{tab:sub-dnn} %  \vskip -0.3cm
\begin{centering}
\begin{tabular}{c|c|c}
\hline 
 & Sub-DNN1 (Image)  & Sub-DNN2 (Text)\tabularnewline
\hline 
Input  & 4096  & 300\tabularnewline
1st layer  & 1024, tanh  & 1024, tanh\tabularnewline
2nd layer  & 1024, tanh  & 1024, tanh\tabularnewline
3rd layer (output)  & 10, linear  & 10, linear\tabularnewline
\hline 
\end{tabular}
\par\end{centering}
\vspace{-1mm}

%\vskip -0.4cm
\end{table}

Original visual feature is denoted as $\boldsymbol{x}\in R^{4096}$
and original textual feature is denoted as $\boldsymbol{y}\in R^{300}$.
The overall functions of sub-DNNs are denoted as $\varphi_{\boldsymbol{X}}=\varphi_{x}(\boldsymbol{x})=f_{3}(\boldsymbol{\Psi}_{3}\cdot f_{2}(\boldsymbol{\Psi}_{2}\cdot f_{1}(\boldsymbol{\Psi}_{1}\boldsymbol{x}+\boldsymbol{b}_{1})+\boldsymbol{b}_{2})+\boldsymbol{b}_{3})$
and $\varphi_{\boldsymbol{Y}}=\varphi_{y}(\boldsymbol{y})$ in a similar
way.

Assume the $i$th pair of samples after sub-DNNs are $\varphi_{\boldsymbol{X}}^{(i)}$
and $\varphi_{\boldsymbol{Y}}^{(i)}$ respectively. The covariance
matrices $\boldsymbol{C}_{xx}$ and $\boldsymbol{C}_{yy}$ are computed
in the same way as in previous works, after removing the average of
$\varphi_{\boldsymbol{X}}^{(i)}$ and $\varphi_{\boldsymbol{Y}}^{(i)}$
and adding a regularization parameter $r$ ($r>0$). Here $E(\cdot)$
is the operation of computing the average. 
\begin{equation}
\boldsymbol{C}_{xx}=E_{i}(\varphi_{\boldsymbol{X}}^{(i)}\varphi_{\boldsymbol{X}}^{(i)T})+r\boldsymbol{I},
\end{equation}

\begin{equation}
\boldsymbol{C}_{yy}=E_{i}(\varphi_{\boldsymbol{Y}}^{(i)}\varphi_{\boldsymbol{Y}}^{(i)T})+r\boldsymbol{I}.
\end{equation}

The cross covariance $\boldsymbol{C}_{xy}$ is computed in a way different
from that of DCCA. It is computed per group (category) $\boldsymbol{g}$
first. Within each group $\boldsymbol{g}$ with the same category,
the general method is to compute $\boldsymbol{C}_{xy}^{(1)}(\boldsymbol{g})$,
using the pairs of visual and textual features from the same venue.
To enhance the similarity within the same group, here we also compute
$\boldsymbol{C}_{xy}^{(2)}(\boldsymbol{g})$, using visual/textual
features from different venues of the same group. Finally, $\boldsymbol{C}_{xy}$
is computed as a weighted average of $\boldsymbol{C}_{xy}^{(1)}(\boldsymbol{g})$
and $\boldsymbol{C}_{xy}^{(2)}(\boldsymbol{g})$ from all data $\boldsymbol{G}$,
using a parameter $\beta$ ($0\le\beta\le1$).

\begin{equation}
\boldsymbol{C}_{xy}^{(1)}(\boldsymbol{g})=E_{i\in\boldsymbol{g}}(\varphi_{\boldsymbol{X}}^{(i)}\varphi_{\boldsymbol{Y}}^{(i)T}),
\end{equation}

\begin{equation}
\boldsymbol{C}_{xy}^{(2)}(\boldsymbol{g})=E_{i,j\in\boldsymbol{g},i\neq j}(\varphi_{\boldsymbol{X}}^{(i)}\varphi_{\boldsymbol{Y}}^{(j)T}),
\end{equation}

\begin{equation}
\boldsymbol{C}_{xy}=\beta\cdot E_{\boldsymbol{g}\subset\boldsymbol{G}}(\boldsymbol{C}_{xy}^{(1)}(\boldsymbol{g}))+(1-\beta)\cdot E_{\boldsymbol{g}\subset\boldsymbol{G}}(\boldsymbol{C}_{xy}^{(2)}(\boldsymbol{g})).\label{eq:var-combi}
\end{equation}
$\beta$ plays an important role here. A large $\beta$ improves the
pairwise correlation but degrades the category-based correlation (C-DCCA
degenerates to DCCA at $\beta=1$). On the other hand, a small $\beta$
improves the category-based correlation but degrades the pairwise
correlation. The two are conflicting targets that cannot be achieved
simultaneously. $\beta$ is set empirically to take a tradeoff between
the two targets. 

In the training phase (part (i) of Fig.~\ref{fig:framework}), $tr(\boldsymbol{W}_{x}^{T}\boldsymbol{C}_{xy}\boldsymbol{W}_{y})$
is computed as an objective function, which is maximized by adjusting
the weights and bias in sub-DNNs. In this process, non-linear functions
$\varphi_{x}(\cdot)$, $\varphi_{y}(\cdot)$, and the linear projections
$\boldsymbol{W}_{x}$ and $\boldsymbol{W}_{y}$ are jointly obtained
as follows

\begin{eqnarray}
(\boldsymbol{W}_{x},\negthinspace\boldsymbol{W}_{y},\negthinspace\varphi_{x},\negthinspace\varphi_{y}) & \negthinspace\negthinspace\negthinspace\negthinspace\negthinspace=\negthinspace\negthinspace\negthinspace\negthinspace\negthinspace\negthinspace\negthinspace\negthinspace\negthinspace\negthinspace & \underset{(\boldsymbol{W}_{x},\boldsymbol{W}_{y},\varphi_{x},\varphi_{y})}{argmax}tr(\boldsymbol{W}_{x}^{T}\boldsymbol{C}_{xy}\boldsymbol{W}_{y}),\nonumber \\
\text{{subject\,to}}: & \negthinspace\negthinspace\negthinspace\negthinspace\negthinspace\negthinspace\negthinspace & \negthinspace\negthinspace\negthinspace\negthinspace\boldsymbol{W}_{x}^{T}\boldsymbol{C}_{xx}\boldsymbol{W}_{x}\negthinspace=\negthinspace\boldsymbol{I,}\boldsymbol{W}_{y}^{T}\boldsymbol{C}_{yy}\boldsymbol{W}_{y}\negthinspace=\negthinspace\boldsymbol{I}.\label{eq:CCA-clust}
\end{eqnarray}

In this optimization, there are four parameters. These parameters
are iteratively solved as follows: 
\begin{enumerate}
\item $\boldsymbol{W}_{x},\boldsymbol{W}_{y},\varphi_{x}$ and $\varphi_{y}$
are initialized. 
\item With a batch of input ($\boldsymbol{X}$, $\boldsymbol{Y}$), their
non-linear mapping result $\varphi_{\boldsymbol{X}}$ and $\varphi_{\boldsymbol{Y}}$
are computed by the sub-DNNs.
\item $\boldsymbol{C}_{xx}$, $\boldsymbol{C}_{yy}$, and $\boldsymbol{C}_{xy}$
are computed.
\item $\boldsymbol{W}_{x}$ and $\boldsymbol{W}_{y}$ are computed by the
CCA.
\item $\varphi_{x}$ and $\varphi_{y}$ are updated by the back propagation
procedure, in the same way as in \cite{Andrew13}.
\item Step 2) to 5) are repeated until the training converges.
\end{enumerate}

\begin{figure}
\centering

\includegraphics[width=7cm]{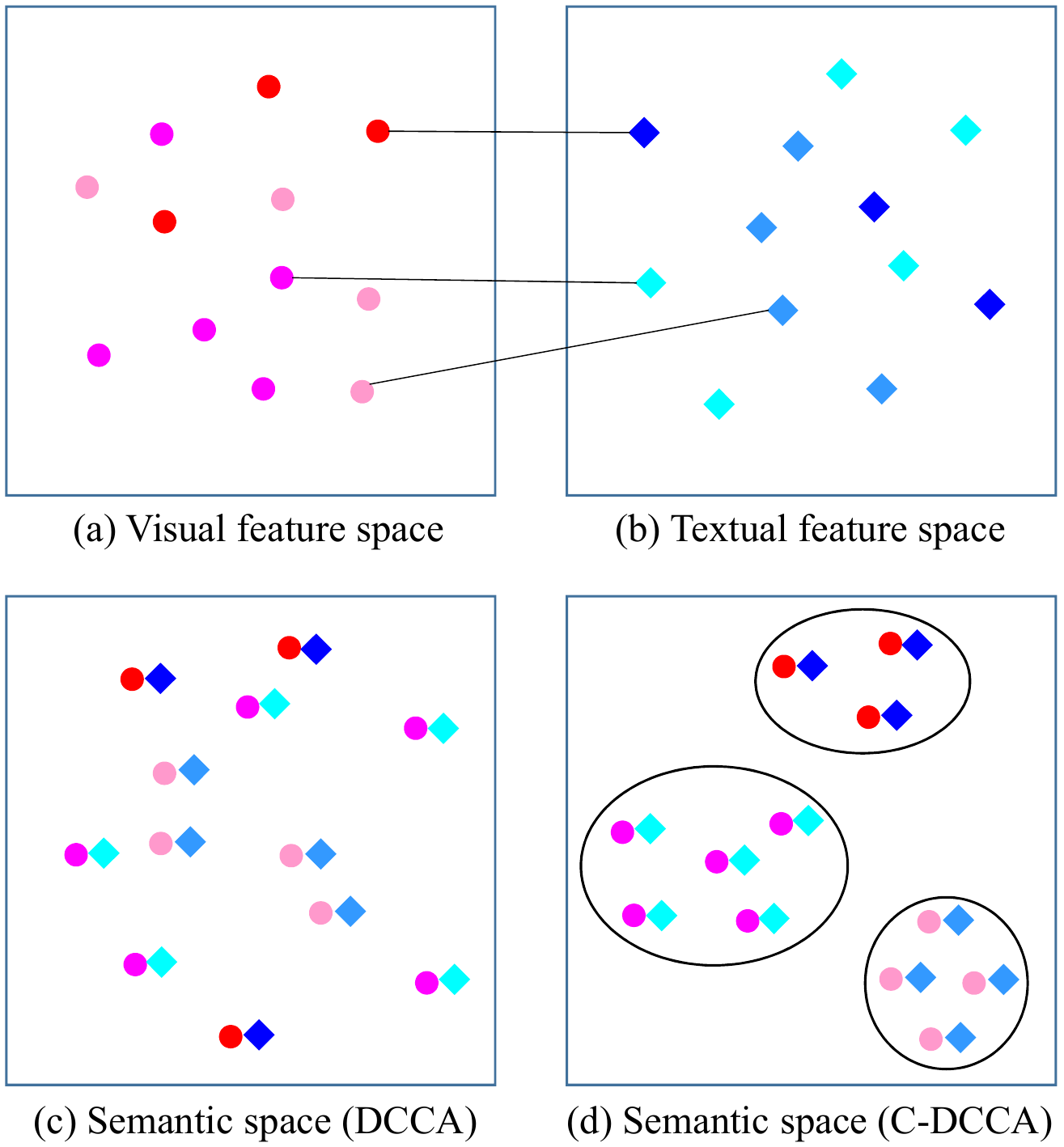}

\vspace{-2mm}

\caption{\label{fig:effect}Expected effect of category-based DCCA. (a) Visual
features and (b) textual features are located in different spaces
with unclear correlation. (c) Visual and textual features are mapped
to the same semantic space by DCCA, strengthening their pairwise correlation.
(d) Besides achieving pairwise correlation, category-based correlation
in C-DCCA makes features from different venues with the same category
close to each other.}

\vspace{-1mm}
\end{figure}

The expected effect of the C-DCCA is shown in Fig.~\ref{fig:effect}.
In Fig.~\ref{fig:effect}(a-b), visual features and textual features
are located in different spaces and have unclear correlations. In
Fig.~\ref{fig:effect}(c), they are mapped to the same semantic space,
strengthening their pairwise correlation. As a comparison, in Fig.~\ref{fig:effect}(d),
category-based correlation in C-DCCA makes features from different
venues with the same category close to each other.

\section{Evaluation}

\label{sec:Evaluation}We evaluate the performance of the proposed
C-DCCA method, by comparing it with CCA \cite{Hotelling36}, KCCA
\cite{Cristianini00}, DCCA \cite{Andrew13}, C-CCA (category-based
CCA) \cite{Rasiwasia14}, and C-KCCA (category-based KCCA) \cite{Rasiwasia14}.
In the evaluation, KCCA uses a Gaussian kernel, and DCCA is based
on \cite{Andrew13}. On top of DCCA, we implemented the proposed C-DCCA.
DCCA shares the same network structure as C-DCCA except that its parameter
$\beta$ is set to 1. The category-based correlation in C-CCA and
C-KCCA is the same as in C-DCCA, with the same parameter $\beta$.
All methods share the same feature (visual feature, text feature,
geographic category), and use the same regularization parameter $r$
and the same number of canonical components (10). 

\subsection{Evaluation setup}

We investigate the fine-grained venue discovery as shown in part (ii)
of Fig.~\ref{fig:framework}. The database is composed of venues
each with a featured article (text description) and other related
information. Given a photo as an input, the system computes the correlation
between the input photo and the text description of venues in the
database. The venue with the maximal correlation is regarded as the
exact venue where the photo was taken.  

In the evaluation, we will use recall-precision curve, mean average
precision (MAP), and mean reciprocal rank 1 (MRR1) as main metrics.
In the group venue search, the system generates a ranked list of venues,
where venues with the same category as that of the input photo are
regarded as relevant. Here, group venue search based on the correlation
between images and text descriptions is different from the one that
merely exploits visual information. It is evaluated by recall-precision
and MAP. When considering whether a specific venue can be discovered
by a photo taken there, there is only one relevant venue, and MRR1
is a good metric indicating both the accuracy and the rank of the
results.

Through integrating venue data including images and texts in Wikipedia
and Foursquare, we have 19,792 photos and 1,994 article descriptions
for 1,994 venues in our experiments. Wikipedia data are collected
from 5 cities (New York City, Los Angeles, London, Sydney, Orlando),
and Foursquare photos are collected from two cities (Los Angeles,
London). Venues with the same name and coordinates in Wikipedia and
Foursquare are regarded as the same. According to the semantics of
venues, Foursquare provides 10 primary categories for venues \cite{Yu14}.
These venue categories (Arts \& Entertainment, College \& University,
Event, Food, Nightlife Spot, Outdoor \& Recreation, Professional \&
Other place, Residence, Shop \& Service, Travel \& Transport) used
for dividing venue photos and descriptions into groups are indexed
from 1 to 10 in our experiment. Category information of venues (images
and texts) is directly obtained from Foursquare by public API. 

Out of the 1,994 Wikipedia venues, 1,500 are used for training the
network. In addition, part of Foursquare images (from both Los Angeles
and London) are added into the training set, paired with Wikipedia
text of the corresponding venue. In the test phase, the database only
consists of textual features of all venues and the queries come from
the Foursquare images (either Los Angeles or London) that are not
used in the training. In each run, the split of Wikipedia dataset
and Foursquare dataset is random for the training, and a new model
is trained. All results are averaged over 5 cross validations.

The experiments are conducted on a Centos7.2 server, which contains
CUDA8.0, Conda3-4.3.11 (python 3.5), Tensorflow 1.3.0, and Keras 2.0.5.
In addition, it is configured with E5-2620v4 CPU (2.1GHz), GeForce
GTX 1080 Ti (11GB), and DDR4-2400 Memory (128GB).

\subsection{Impact of Different Parameters}

In the training, the Adam optimizer is used and the learning rate
is set to 0.0001. Batch size is set to 100. We tried different regularization
parameter $r$ (0.01, 0.001, 0.0001, 0.00001), but found no significant
difference. In the rest experiments, $r$ is set to 0.0001.  

\begin{figure}
\centering

\includegraphics[width=7.5cm]{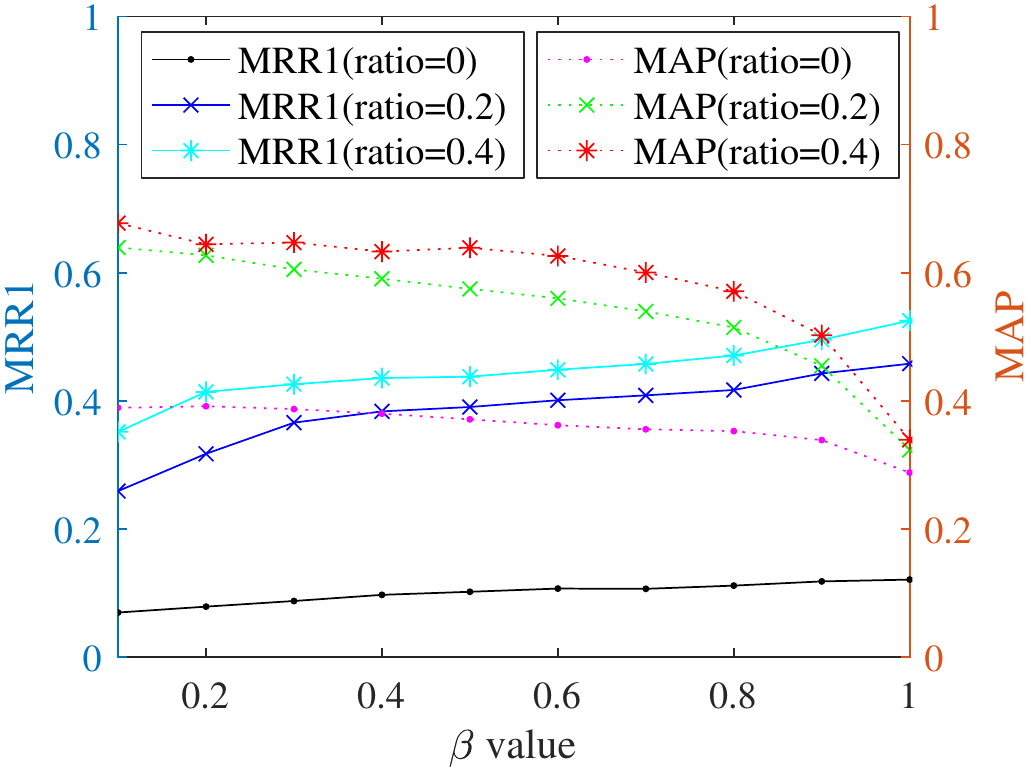}

\vspace{-2mm}
\caption{\label{fig:param-beta} MRR1 and MAP under different values of $\beta$
(Using Los Angeles photos as query, the ratio of Foursquare photos
used in the training is equal to 0, 0.2 or 0.4).}

\vspace{-1mm}
\end{figure}

The parameter $\beta$ greatly affects system performance. The results
of MRR1 and MAP, under different values of $\beta$, are shown in
Fig.~\ref{fig:param-beta}. As $\beta$ increases, MRR1 increases
while MAP decreases. This is because a large $\beta$ will lead to
a large weight for the covariance and a small weight for the cross
covariance in Eq.~\ref{eq:var-combi}. There is no significant variation
in the performance when $\beta$ changes within the range of $(0.3,0.7)$.
In the following $\beta$ is set to 0.3.

\subsection{Evaluation of the Group Venue Search}

Here we evaluate the group venue search. We adjust the ratio of Foursquare
photos in the training to see how well all methods benefit from the
increase of photos in each venue.

\begin{figure}
\centering

\includegraphics[width=7.5cm]{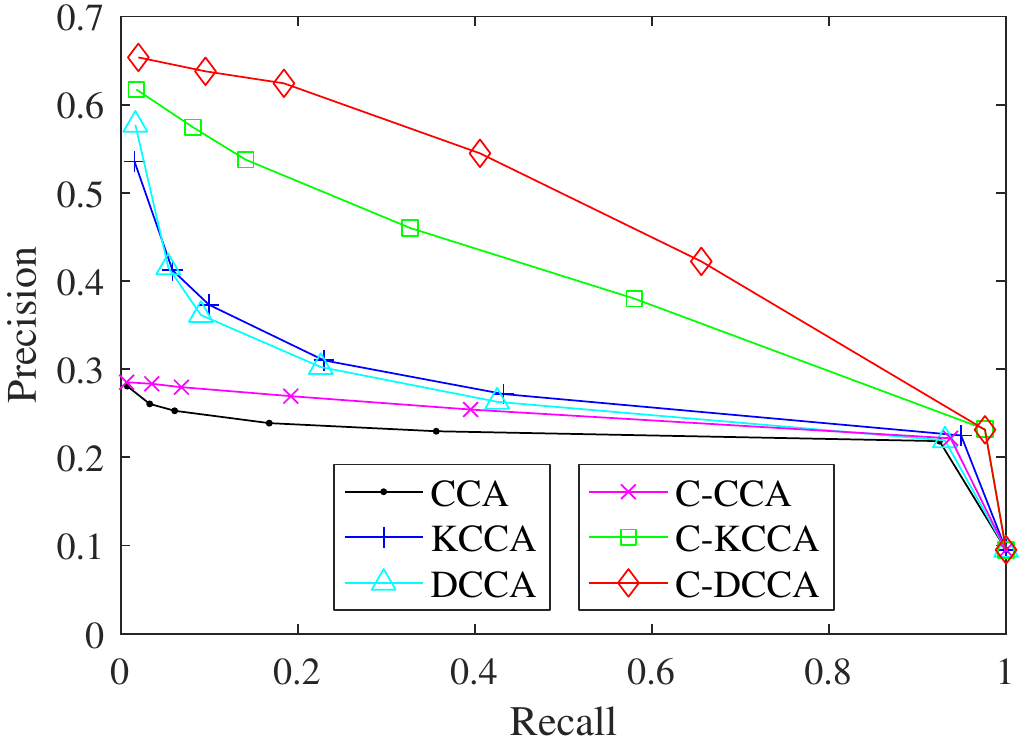}

\vspace{-2mm}
\caption{\label{fig:recall_prec_LA} Recall-precision relationship, obtained
by changing the number of output (using Los Angeles photos in Foursquare
as queries, ratio of Foursquare photos for training = 20\%).}

\vspace{-1mm}
\end{figure}

Fig.~\ref{fig:recall_prec_LA} demonstrates the recall-precision
curve using Los Angeles photos as queries, where the ratio of Foursquare
photos used in the training is equal to 20\%. This pair of recall
and precision is obtained by changing the number of output. Generally,
C-DCCA outperforms C-KCCA, which outperforms other methods. As the
number of output increases, recall gradually increases while precision
decreases, but with different trends in different methods. The decrease
in precision is slow in C-DCCA and C-KCCA which indicates that most
venues in the top have the same category as the input photo, but the
decrease in precision is fast in DCCA and KCCA, which implies that
except the first venue, other venues may have different categories
as the input photo. This confirms the fact that C-DCCA improves the
category-based correlation while DCCA only stresses the pairwise correlation. 

\begin{figure}[tb]
\centering

\includegraphics[width=7.5cm]{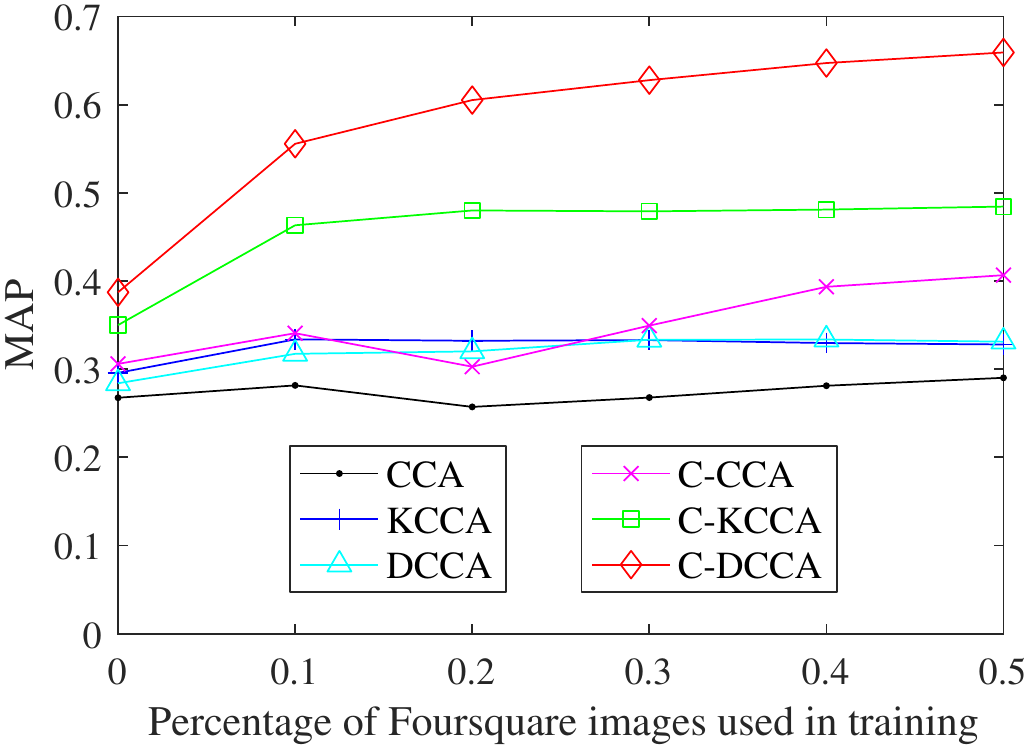}

\vspace{-2mm}
\caption{\label{fig:map_LA} Mean average precision in the group venue search
(using Los Angeles photos in Foursquare as queries). }

\vspace{-1mm}
\end{figure}

The MAP result is shown in Fig.~\ref{fig:map_LA}. By increasing
the percentage of Foursquare photos in the training, more photos are
used to represent venues in learning the correlation between photos
and text descriptions. Therefore, the textual feature can better represent
a venue. Accordingly, the MAP result usually increases in all methods,
but with different trends. MAP in C-DCCA has a much larger gain than
that of the other methods.

\begin{figure}[tb]
\centering

\includegraphics[width=7.5cm]{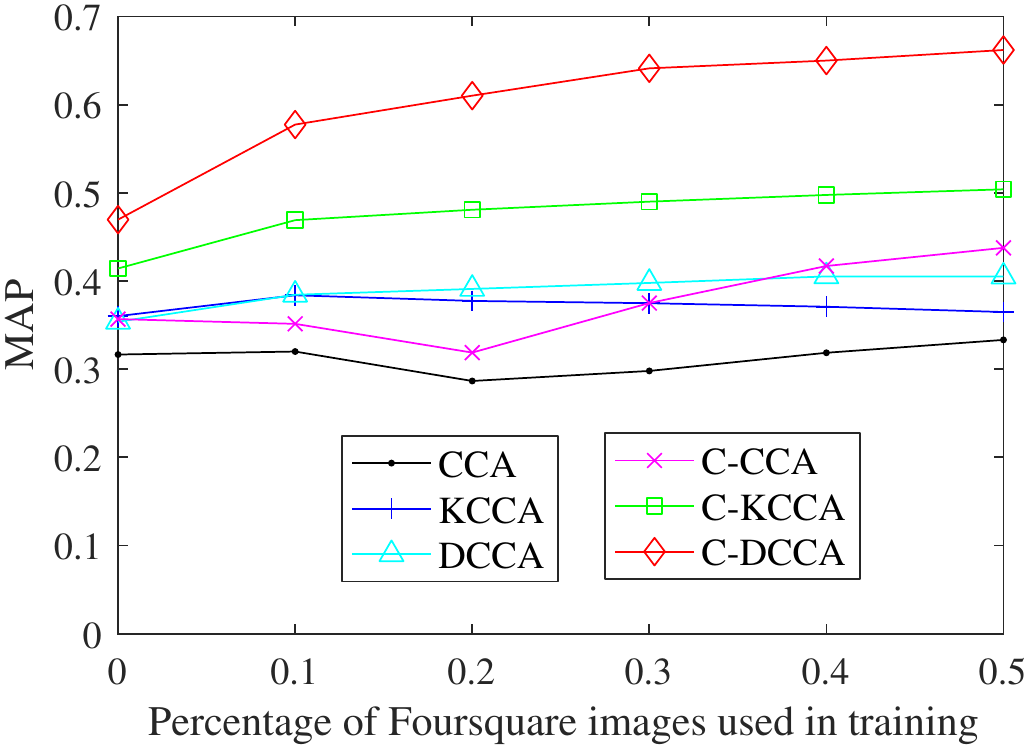}

\vspace{-2mm}
\caption{\label{fig:map_London} Mean average precision in the group venue
search (using London photos in Foursquare as queries). }

\vspace{-2mm}
\end{figure}

\begin{figure}[tb]
\centering

\includegraphics[width=7.5cm]{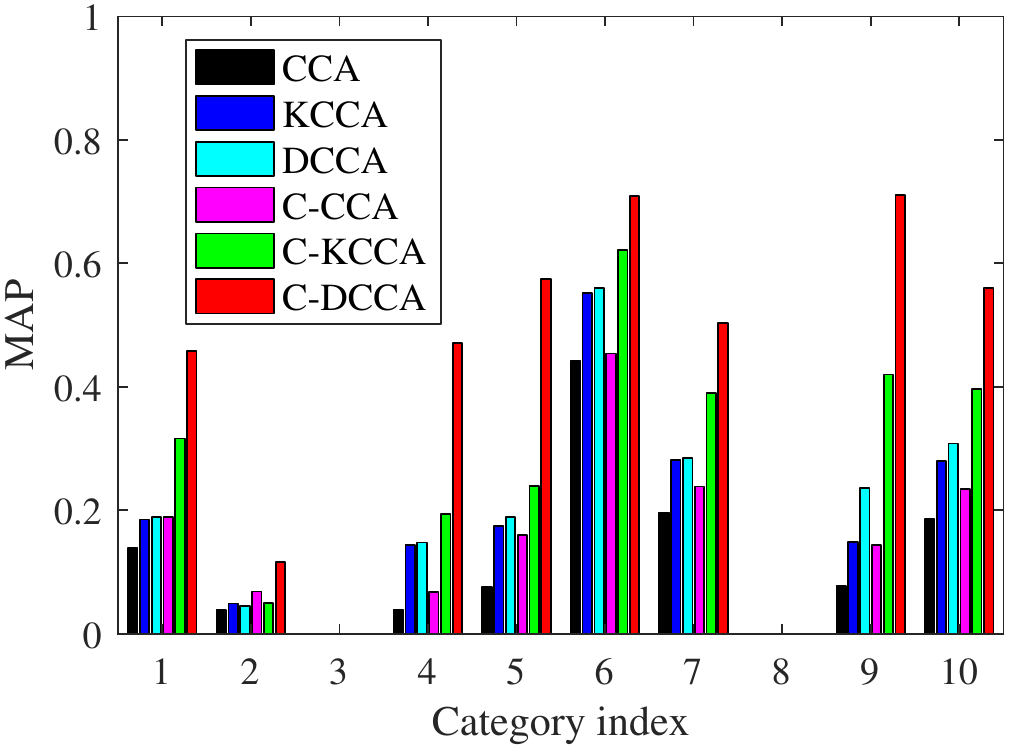}

\vspace{-2mm}
\caption{\label{fig:map_percat_London} Mean average precision per category
in the group venue search (using London photos in Foursquare as queries,
ratio of Foursquare photos for training = 20\%).}
\vspace{-2mm}
\end{figure}

Next we use London photos as queries, which has many more photos than
Los Angeles. The MAP result in Fig.~\ref{fig:map_London} shows a
similar trend as in Fig.~\ref{fig:map_LA}. Fig.~\ref{fig:map_percat_London}
further shows the MAP per category. There are no venues with 3rd and
8th categories. In other categories, C-DCCA achieves much better performance
than other methods. 

\subsection{Evaluation of the Exact Venue Search}

\begin{figure}[tb]
\centering

\includegraphics[width=7.5cm]{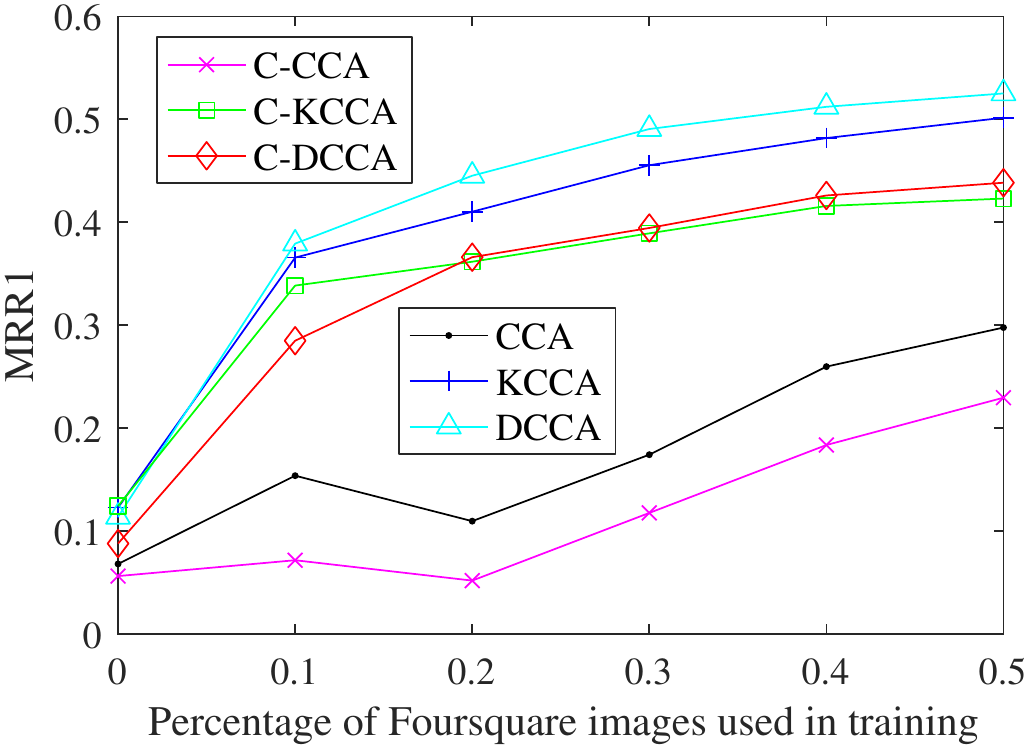}

\vspace{-2mm}
\caption{\label{fig:mrr1_LA} MRR1 in the exact venue search (Los Angeles data
in Foursquare). }

\vspace{-1mm}
\end{figure}

The performance of finding the exact venue where a photo was taken
is evaluated by the MRR1 metric. The MRR1 result with Los Angeles
photos in Foursquare as queries is shown in Fig.~\ref{fig:mrr1_LA}.
Although the MRR1 performance increases with the ratio of Foursquare
images used in the training in all methods, there is a clear gap between
C-DCCA and DCCA. But compared with the decrease of MRR1 in C-DCCA,
the increase of MAP in C-DCCA is much larger.

In the above, we only considered the correlation between the photo
and text. Actually, when a photo was taken by a mobile device, the
position of the photo can somehow be obtained simultaneously and used
to narrow the search range. A coarse positioning method that works
in both outdoor and indoor environments is to collect the signal strength
and cell ID of nearby cells. Although this accuracy is very coarse,
it does help to reduce the search range and improve the performance
of finding relevant venues. 

\begin{figure}[tb]
\centering

\includegraphics[width=7.5cm]{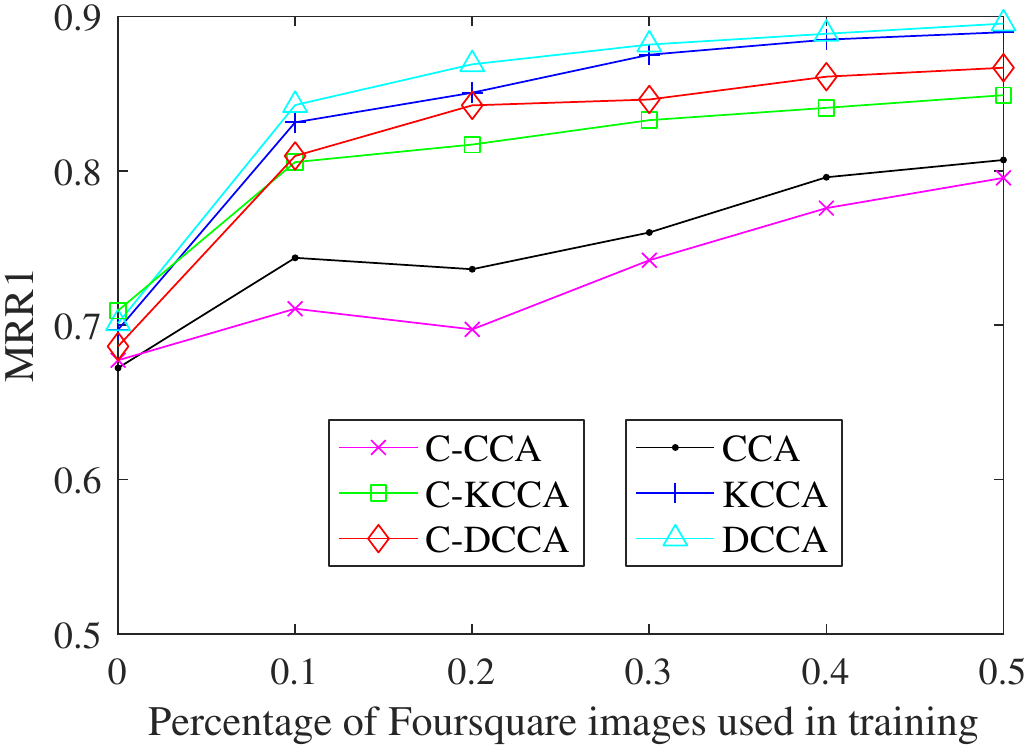}

\vspace{-2mm}
\caption{\label{fig:mrr1_LA_1km} MRR1 in the exact venue search by considering
coarse position information (using Los Angeles photos in Foursquare
as queries). }

\vspace{-1mm}
\end{figure}

\begin{figure}[tb]
\centering

\includegraphics[width=7.5cm]{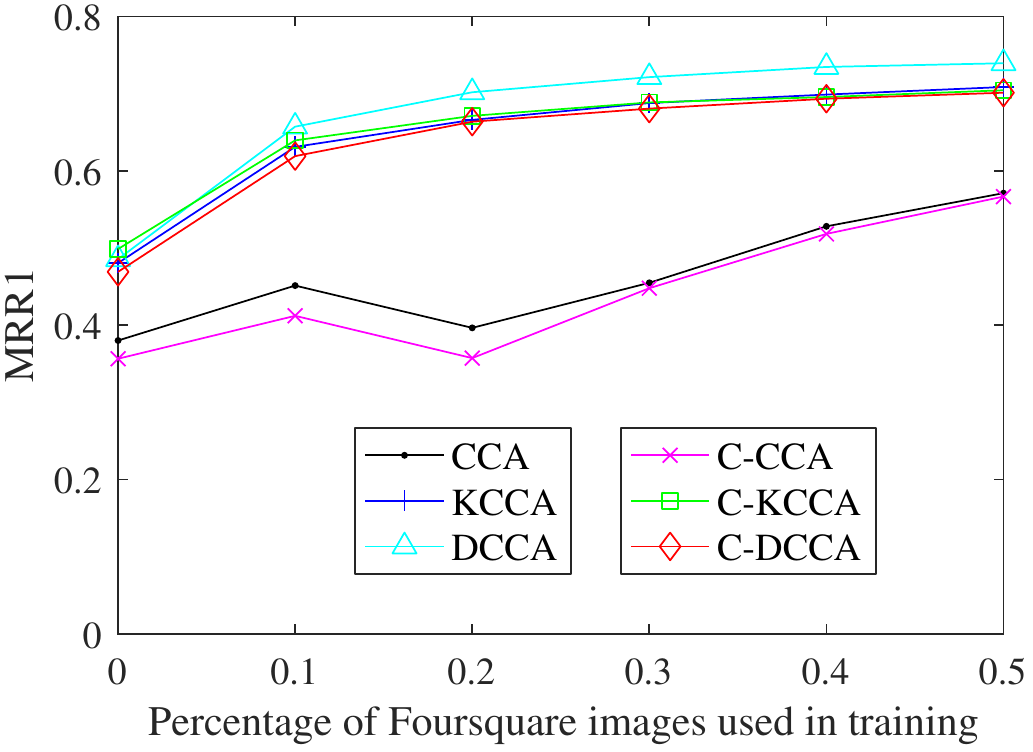}

\vspace{-2mm}
\caption{\label{fig:mrr1_London_1km} MRR1 in the exact venue search by considering
coarse position information (using London photos in Foursquare as
queries). }

\vspace{-1mm}
\end{figure}

Assume that the position of each venue is accurately known and that
the positioning accuracy of a photo by using cell ID is 1km. Then,
the MRR1 results, using Los Angeles photos or London photos as queries,
are shown in Fig.~\ref{fig:mrr1_LA_1km} and Fig.~\ref{fig:mrr1_London_1km},
respectively. The gap between C-DCCA and DCCA decreases. In Los Angeles,
MRR1 is above 0.8. In London, MRR1 is a little lower due to more venues
there, but it is still nearly 0.7.

In this way, C-DCCA not only has nearly the same performance (MRR1)
as DCCA in the exact venue search, but also achieves much better performance
(MAP) in the group venue search compared with DCCA and other methods. 

\subsection{Evaluation with the UCSD Dataset}

To demonstrate its applicability and effectiveness in other tasks,
we extend our method to the cross-modal retrieval and classification
between image and text over a public dataset provided by the UCSD
group \cite{Rasiwasia10}. This multimodal image-text dataset is completely
generated from Wikipedia featured articles without relying on other
image resources, which contains 2,866 documents (each contains a pair
of text and image) in 10 most popular categories. It is divided into
a training set of 2,173 documents and a testing set of 693 documents.
The definition of category depends on Wikipedia tags and is different
from that of Foursquare with venue-specific semantic tags.

In \cite{Rasiwasia10}, the textual feature is derived from a LDA
model with 10 dimensions. An image is first represented as a bag of
SIFT descriptions, and visual feature space is divided into 128 sub-spaces
by k-means. Then, a new visual feature is computed as the SIFT histograms
(SIFThist) with respect to these sub-spaces, which has 128 dimensions.
The results of CM (correlation matching), SM (semantic matching) and
SCM (semantic correlation matching) are directly taken from \cite{Rasiwasia10}.

We explore the different combinations of visual feature (VGG16 or
SIFThist) and textual feature (Doc2Vec or LDA) to evaluate the performance
of C-DCCA. For VGG16+Doc2Vec, the same network architecture in Table~\ref{tab:sub-dnn}
is used. For other combinations, the input is changed accordingly,
and the number of CCA component is adjusted a little to reach their
best performance. We also try to fine tune the pretrained models.
With the training set and the category information, we refine the
fully-connected layers of VGG16. Because the Doc2Vec model does not
support fine tuning with a different vocabulary set, we retrain it
from the ground.

The MAP results of group venue search are shown in Table~\ref{tab:UCSD-result}.
These results reflect three facts. (i) C-DCCA outperforms the cross-modal
retrieval method suggested in \cite{Rasiwasia10} in terms of correlation
analysis. Compared with CM/SCM, C-DCCA achieves a better performance
when using the same hand-crafted SIFThist+LDA feature. (ii) Features
also play an important role. Because VGG16 far outperforms SIFThist
and Doc2Vec is also a little better than LDA, we can see a large improvement
when C\textendash DCCA changes its feature from SIFThist+LDA to VGG16+Doc2Vec.
(iii) Refining the pretrained model is also helpful. The re-training
of the pre-trained models has an obvious effect (especially when VGG16+Doc2Vec
is used) because the statistical property of the UCSD dataset is not
completely the same as that of large datasets used for pretraining
these models.

\begin{table}[!t]
\caption{MAP result on the UCSD dataset (image query)}

\label{tab:UCSD-result} %  \vskip -0.3cm
\begin{centering}
\begin{tabular}{c|c|c|c|c}
\hline 
Visual feature & VGG16  & SIFThist & VGG16 & SIFThist\tabularnewline
Textual feature & Doc2Vec & Doc2Vec & LDA & LDA\tabularnewline
C-DCCA & 0.455 & 0.270 & 0.448 & 0.310\tabularnewline
C-DCCA (re-train) & 0.496 & 0.269 & 0.466 & N/A\tabularnewline
CM \cite{Rasiwasia10}{]} & N/A & N/A & N/A & 0.249\tabularnewline
SM \cite{Rasiwasia10} & N/A & N/A & N/A & 0.225\tabularnewline
SCM \cite{Rasiwasia10} & N/A & N/A & N/A & 0.277\tabularnewline
\hline 
\end{tabular}
\par\end{centering}
\vspace{-1mm}

%\vskip -0.4cm
\end{table}

\section{Conclusion}

\label{sec:Conclusion}

This paper studies fine-grained venue discovery by learning the cross-modal
correlation between photos and the rich text descriptions of venues
over heterogeneous multimodal contents. Different from previous research,
this work jointly optimizes the pairwise correlation and the category-based
correlation, and the exact venue search and group venue search are
realized simultaneously. Extensive experiments confirm that (i) The
proposed C-DCCA method greatly improves the performance of group venue
discovery, compared to state-of-the-art methods. (ii) Using the coarse
location information helps to reduce the gap between C-DCCA and DCCA
in the exact venue search. (iii) Using extra image resources for representing
visual aspects of venues helps to further improve the performance
of fine-grained venue discovery.

\bibliographystyle{IEEEtran}
\bibliography{IEEEabrv,mybibfile}

\begin{thebibliography}{10}
\providecommand{\url}[1]{#1}
\csname url@samestyle\endcsname
\providecommand{\newblock}{\relax}
\providecommand{\bibinfo}[2]{#2}
\providecommand{\BIBentrySTDinterwordspacing}{\spaceskip=0pt\relax}
\providecommand{\BIBentryALTinterwordstretchfactor}{4}
\providecommand{\BIBentryALTinterwordspacing}{\spaceskip=\fontdimen2\font plus
\BIBentryALTinterwordstretchfactor\fontdimen3\font minus
  \fontdimen4\font\relax}
\providecommand{\BIBforeignlanguage}[2]{{%
\expandafter\ifx\csname l@#1\endcsname\relax
\typeout{** WARNING: IEEEtran.bst: No hyphenation pattern has been}%
\typeout{** loaded for the language `#1'. Using the pattern for}%
\typeout{** the default language instead.}%
\else
\language=\csname l@#1\endcsname
\fi
#2}}
\providecommand{\BIBdecl}{\relax}
\BIBdecl

\bibitem{Hays08}
J.~Hays and A.~A. Efros, ``{IM2GPS:} estimating geographic information from a
  single image,'' in \emph{2008 {IEEE} Computer Society Conference on Computer
  Vision and Pattern Recognition, {CVPR}}, 2008.

\bibitem{Chen16}
B.-C. Chen, Y.-Y. Chen, F.~Chen, and D.~Joshi, ``Business-aware visual concept
  discovery from social media for multimodal business venue recognition,'' in
  \emph{Proceedings of the Thirtieth AAAI Conference on Artificial
  Intelligence}, ser. AAAI'16, pp. 101--107.

\bibitem{Schindler07}
G.~Schindler, M.~A. Brown, and R.~Szeliski, ``City-scale location
  recognition,'' in \emph{2007 {IEEE} Computer Society Conference on Computer
  Vision and Pattern Recognition {CVPR}}, 2007.

\bibitem{Zhou14}
B.~Zhou, A.~Lapedriza, J.~Xiao, A.~Torralba, and A.~Oliva, ``Learning deep
  features for scene recognition using places database,'' in \emph{Proceedings
  of the 27th International Conference on Neural Information Processing
  Systems}, ser. NIPS'14, 2014, pp. 487--495.

\bibitem{Simonyan14}
K.~Simonyan and A.~Zisserman, ``Very deep convolutional networks for
  large-scale image recognition,'' \emph{CoRR}, vol. abs/1409.1556, 2014.

\bibitem{Szegedy15}
C.~Szegedy, W.~Liu, Y.~Jia, P.~Sermanet, S.~Reed, D.~Anguelov, D.~Erhan,
  V.~Vanhoucke, and A.~Rabinovich, ``Going deeper with convolutions,'' in
  \emph{2015 IEEE Conference on Computer Vision and Pattern Recognition
  (CVPR)}, 2015, pp. 1--9.

\bibitem{Donahue14}
J.~Donahue, Y.~Jia, O.~Vinyals, J.~Hoffman, N.~Zhang, E.~Tzeng, and T.~Darrell,
  ``Decaf: A deep convolutional activation feature for generic visual
  recognition,'' in \emph{Proceedings of the 31st International Conference on
  International Conference on Machine Learning - Volume 32}, ser. ICML'14,
  2014, pp. I--647--I--655.

\bibitem{Rajiv16}
R.~R. Shah, Y.~Yu, A.~Verma, S.~Tang, A.~D. Shaikh, and R.~Zimmermann,
  ``Leveraging multimodal information for event summarization and concept-level
  sentiment analysis,'' \emph{Know.-Based Syst.}, vol. 108, no.~C, pp.
  102--109, 2016.

\bibitem{Malinowski15}
M.~Malinowski, M.~Rohrbach, and M.~Fritz, ``Ask your neurons: {A} neural-based
  approach to answering questions about images,'' in \emph{2015 {IEEE}
  International Conference on Computer Vision, {ICCV}}, 2015, pp. 1--9.

\bibitem{Yan15}
F.~Yan and K.~Mikolajczyk, ``Deep correlation for matching images and text,''
  in \emph{{IEEE} Conference on Computer Vision and Pattern Recognition,
  {CVPR}}, 2015, pp. 3441--3450.

\bibitem{Rasiwasia10}
\BIBentryALTinterwordspacing
N.~Rasiwasia, J.~Costa~Pereira, E.~Coviello, G.~Doyle, G.~R. Lanckriet,
  R.~Levy, and N.~Vasconcelos, ``A new approach to cross-modal multimedia
  retrieval,'' in \emph{Proceedings of the 18th ACM International Conference on
  Multimedia}, ser. MM '10.\hskip 1em plus 0.5em minus 0.4em\relax New York,
  NY, USA: ACM, 2010, pp. 251--260. [Online]. Available:
  \url{http://doi.acm.org/10.1145/1873951.1873987}
\BIBentrySTDinterwordspacing

\bibitem{Rasiwasia14}
N.~Rasiwasia, D.~Mahajan, V.~Mahadevan, and G.~Aggarwal, ``Cluster canonical
  correlation analysis,'' in \emph{Proceedings of International Conference on
  Artificial Intelligence and Statistics}, ser. AISTATS'14, vol.~33, pp.
  823--831.

\bibitem{Andrew13}
G.~Andrew, R.~Arora, J.~Bilmes, and K.~Livescu, ``Deep canonical correlation
  analysis,'' in \emph{Proceedings of the 30th International Conference on
  International Conference on Machine Learning - Volume 28}, ser. ICML'13,
  2013, pp. III--1247--III--1255.

\bibitem{Hotelling36}
H.~Hotelling, ``Relations between two sets of variates,'' \emph{Biometrika},
  vol.~28, no. 3/4, pp. 321--377, 1936.

\bibitem{Li03}
D.~Li, N.~Dimitrova, M.~Li, and I.~K. Sethi, ``Multimedia content processing
  through cross-modal association,'' in \emph{Proceedings of the Eleventh ACM
  International Conference on Multimedia}, ser. MM '03.\hskip 1em plus 0.5em
  minus 0.4em\relax ACM, 2003, pp. 604--611.

\bibitem{Tenenbaum00}
J.~B. Tenenbaum and W.~T. Freeman, ``Separating style and content with bilinear
  models,'' \emph{Neural Comput.}, vol.~12, no.~6, pp. 1247--1283, Jun. 2000.

\bibitem{Cristianini00}
N.~Cristianini and J.~Shawe-Taylor, \emph{An Introduction to Support Vector
  Machines: And Other Kernel-based Learning Methods}.\hskip 1em plus 0.5em
  minus 0.4em\relax Cambridge University Press, 2000.

\bibitem{Friedland10}
G.~Friedland, O.~Vinyals, and T.~Darrell, ``Multimodal location estimation,''
  in \emph{Proceedings of the 18th ACM International Conference on Multimedia},
  ser. MM '10, 2010, pp. 1245--1252.

\bibitem{Li09}
Y.~Li, D.~J. Crandall, and D.~P. Huttenlocher, ``Landmark classification in
  large-scale image collections,'' in \emph{{IEEE} 12th International
  Conference on Computer Vision, {ICCV}}, 2009, pp. 1957--1964.

\bibitem{Chen11}
D.~M. Chen, G.~Baatz, K.~Koser, S.~S. Tsai, R.~Vedantham, T.~Pylvanainen,
  K.~Roimela, C.~Xin, J.~Bach, M.~Pollefeys, B.~Girod, and R.~Grzeszczuk,
  ``City-scale landmark identification on mobile devices,'' in
  \emph{Proceedings of the 2011 IEEE Conference on Computer Vision and Pattern
  Recognition}, ser. CVPR '11, 2011, pp. 737--744.

\bibitem{Lin15}
T.~Lin, Y.~Cui, S.~J. Belongie, and J.~Hays, ``Learning deep representations
  for ground-to-aerial geolocalization,'' in \emph{{IEEE} Conference on
  Computer Vision and Pattern Recognition, {CVPR}}, 2015, pp. 5007--5015.

\bibitem{Dalal05}
N.~Dalal and B.~Triggs, ``Histograms of oriented gradients for human
  detection,'' in \emph{2005 IEEE Computer Society Conference on Computer
  Vision and Pattern Recognition (CVPR'05)}, vol.~1, 2005, pp. 886--893 vol. 1.

\bibitem{Lowe99}
D.~G. Lowe, ``Object recognition from local scale-invariant features,'' in
  \emph{Proceedings of the International Conference on Computer Vision-Volume 2
  - Volume 2}, ser. ICCV '99, 1999, p. 1150.

\bibitem{Lau16}
J.~H. Lau and T.~Baldwin, ``An empirical evaluation of doc2vec with practical
  insights into document embedding generation,'' \emph{CoRR}, vol.
  abs/1607.05368, 2016.

\bibitem{ManningSBFBM14}
C.~D. Manning, M.~Surdeanu, J.~Bauer, J.~R. Finkel, S.~Bethard, and
  D.~McClosky, ``The stanford corenlp natural language processing toolkit,'' in
  \emph{Proceedings of the 52nd Annual Meeting of the Association for
  Computational Linguistics, {ACL} 2014}, 2014, pp. 55--60.

\bibitem{Yu14}
Y.~Yu, S.~Tang, R.~Zimmermann, and K.~Aizawa, ``Empirical observation of user
  activities: Check-ins, venue photos and tips in foursquare,'' in
  \emph{Proceedings of the First International Workshop on Internet-Scale
  Multimedia Management}, ser. WISMM '14, 2014, pp. 31--34.

\end{thebibliography}

\end{document}